# A Method for Implementing a Probabilistic Model as a Relational Database

S.K.M. Wong, C.J. Butz, and Y. Xiang
Department of Computer Science
University of Regina
Regina, Saskatchewan
Canada, S4S 0A2
< {wong,butz,yxiang}@cs.uregina.ca>

## Abstract

This paper discusses a method for implementing a probabilistic inference system based on an extended relational data model. This model provides a unified approach for a variety of applications such as dynamic programming, solving sparse linear equations, and constraint propagation. In this framework, the probability model is represented as a generalized relational database. Subsequent probabilistic requests can be processed as standard relational queries. Conventional database management systems can be easily adopted for implementing such an approximate reasoning system.

## 1 Introduction

Probabilistic models [4, 9, 10] are used for making decisions under uncertainty. The input to a probabilistic model is usually a Bayesian network [10]. It may also consist of a set of potentials which define a Markov network [4]. In this paper, we assume that the probabilistic model is described by a Markov network. For this model, the propagation method [5, 6, 7, 12, 13] can be conveniently applied to convert the potentials into marginal distributions.

There is another important reason to characterize a probabilistic model by a Markov network, as it has been shown that such a network can be represented as a generalized relational database [14, 15, 16]. That is, the probabilistic model can be transformed into an equivalent (extended) relational data model. More specifically, the marginal corresponding to each potential can be viewed as a relation in the relational database. Furthermore, the database scheme derived from a Markov network forms an acyclic join dependency [15], which possesses many desirable properties [1, 8] in database applications.

As the probabilistic model is now represented by a relational data model, a probability request expressed as a conditional probability can be equivalently transformed into a standard query to be executed by the database management system. Naturally, all query optimization techniques can be directly applied to processing this query including data structure modification. Thus, these transformations allow us to take full advantage of the query optimizer and other performance enhancement capabilities available in traditional relational databases.

This paper, a sequel of the presentation in the IPMU conference [15], reports on the technical details involved in the design of a probabilistic inference system by transforming a Markov network into a relational database.

Our paper is organized as follows. In Section 2, for completeness we review a unified relational data model for both probabilistic reasoning and database management systems. In Section 3, we show that a factored probability distribution can be expressed as a generalized acyclic join dependency. The method for implementing a probabilistic inference system is described in Section 4. First, we describe how a relational database is constructed for a given probabilistic model. We then show that processing a request for evidential reasoning is equivalent to processing a standard relational query. We conclude by pointing out that the extended relational database system can in fact model a number of apparently different but closely related applications [12].

## 2 An Extended Relational Data Model for Probabilistic Inference

Before introducing our data model, we need to define some basic notions pertinent to our discussion such as: hypergraphs, factored distributions, and marginalization. Then we show how under certain conditions a factored joint probability distribution can be expressed as a generalized acyclic join dependency in the extended relational model.

## 2.1 Basic Notions

*Hypergraphs and Hypertrees* :

Let $\mathcal{L}$ denote a lattice. We say that $\mathcal{H}$ is a *hypergraph*, if $\mathcal{H}$ is a finite subset of $\mathcal{L}$. Consider, for example, the power set $2^{\mathcal{X}}$, where $\mathcal{X} = \{x_1, x_2, ..., x_n\}$ is a set of variables. The power set $2^{\mathcal{X}}$ is a lattice of all subsets of $\mathcal{X}$. Any subset of $2^{\mathcal{X}}$ is a hypergraph on $2^{\mathcal{X}}$. We say that an element $t$ in a hypergraph $\mathcal{H}$ is a *twig* if there exists another element $b$ in $\mathcal{H}$, distinct from $t$, such that $t \cap (\cup(\mathcal{H} - \{t\})) = t \cap b$. We call any such $b$ a *branch* for the twig $t$. A hypergraph $\mathcal{H}$ is a *hypertree* (an acyclic hypergraph [1]) if its elements can be ordered, say $h_1, h_2, ..., h_n$, so that $h_i$ is a twig in $\{h_1, h_2, ..., h_i\}$, for $i = 2, ..., n$. We call any such ordering a hypertree *construction ordering* for $\mathcal{H}$. Given a hypertree construction ordering $h_1, h_2, ..., h_n$, we can choose, for $i$ from 2 to $n$, an integer $b(i)$ such that $1 \leq b(i) \leq i - 1$ and $h_{b(i)}$ is a branch for $h_i$ in $\{h_1, h_2, ..., h_i\}$. We call the function $b(i)$ satisfying this condition a branching function for $\mathcal{H}$ and $h_1, h_2, ..., h_n$.

For example, let $\mathcal{X} = \{x_1, x_2, ..., x_6\}$ and $\mathcal{L} = 2^{\mathcal{X}}$. Consider a hypergraph, $\mathcal{H} = \{h_1 = \{x_1, x_2, x_3\}, h_2 = \{x_1, x_2, x_4\}, h_3 = \{x_2, x_3, x_5\}, h_4 = \{x_5, x_6\}\}$, depicted in Figure 1. This hypergraph is in fact a hypertree; the ordering, for example, $h_1, h_2, h_3, h_4$, is a hypertree construction ordering and $b(2) = 1, b(3) = 1$, and $b(4) = 3$ define its branching function.

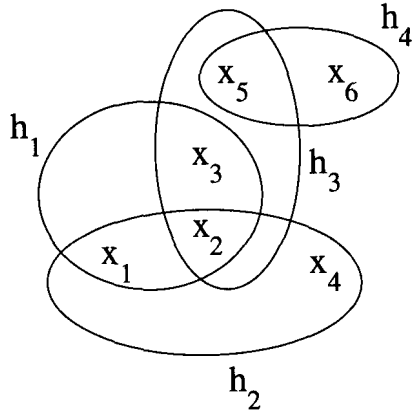

Figure 1: A graphical representation of the hypergraph $\mathcal{H} = \{h_1, h_2, h_3, h_4\}$.

A hypertree $\mathcal{K}$ on $\mathcal{L}$ is called a *hypertree cover* for a given hypergraph $\mathcal{H}$ on $\mathcal{L}$ if for every element $h$ of $\mathcal{H}$, there exists an element $k(h)$ of $\mathcal{K}$ such that $h \subseteq k(h)$. In general, a hypergraph $\mathcal{H}$ may have many hypertree covers. For example, the hypertree depicted in Figure 1 is a hypertree cover of the hypergraph, $\{\{x_1, x_2\}, \{x_1, x_3\}, \{x_1, x_2, x_4\}, \{x_2, x_5\}, \{x_3, x_5\}, \{x_5, x_6\}\}$.

*Factored Probability Distributions* :

Let $\mathcal{X} = \{x_1, x_2, ..., x_n\}$ denote a set of variables. A factored probability distribution $p(x_1, x_2, ..., x_n)$ can be written as:

$$p(x_1, x_2, ..., x_n) = \phi = \phi_{h_1} \phi_{h_2} ... \phi_{h_n},$$

where each $h_i$ is a subset of $\mathcal{X}$, i.e., $h_i \in 2^{\mathcal{X}}$, and $\phi_{h_i}$ is a real-valued function on $h_i$. Moreover, $\mathcal{X} = h_1 \cup h_2 \cup ... \cup h_n = \bigcup_{i=1}^{n} h_i$. By definition, $\mathcal{H} = \{h_1, h_2, ..., h_n\}$ is a hypergraph on the lattice $2^{\mathcal{X}}$. Thus, a factored probability distribution can be viewed as a product on a hypergraph $\mathcal{H}$, namely:

$$\phi = \prod_{h \in \mathcal{H}} \phi_h.$$

Let $v_x$ denote the discrete frame (state space) of the variable $x \in \mathcal{X}$. We call an element of $v_x$ a *configuration* of $x$. We define $v_h$ to be the Cartesian product of the frames of the variables in a hyperedge $h \in 2^{\mathcal{X}}$:

$$v_h = \bigtimes_{x \in h} v_x.$$

We call $v_h$ the frame of $h$, and we call its elements configurations of $h$.

Let $h, k \in 2^{\mathcal{X}}$, and $h \subseteq k$. If $\mathbf{c}$ is a configuration of $k$, i.e., $\mathbf{c} \in v_k$, we write $\mathbf{c}^{\downarrow h}$ for the configuration of $h$ obtained by deleting the values of the variables in $k$ and not in $h$. For example, let $h = \{x_1, x_2\}, k = \{x_1, x_2, x_3, x_4\}$, and $\mathbf{c} = (c_1, c_2, c_3, c_4)$, where $c_i \in v_{x_i}$. Then, $\mathbf{c}^{\downarrow h} = (c_1, c_2)$.

If $h$ and $k$ are disjoint subsets of $\mathcal{X}$, $\mathbf{c}_h$ is a configuration of $h$, and $\mathbf{c}_k$ is a configuration of $k$, then we write $(\mathbf{c}_h \circ \mathbf{c}_k)$ for the configuration of $h \cup k$ obtained by *concatenating* $\mathbf{c}_h$ and $\mathbf{c}_k$. In other words, $(\mathbf{c}_h \circ \mathbf{c}_k)$ is the unique configuration of $h \cup k$ such that $(\mathbf{c}_h \circ \mathbf{c}_k)^{\downarrow h} = \mathbf{c}_h$ and $(\mathbf{c}_h \circ \mathbf{c}_k)^{\downarrow k} = \mathbf{c}_k$. Using the above notation, a factored probability distribution $\phi$ on $\cup \mathcal{H}$ can be defined as follows:

$$\phi(\mathbf{c}) = (\prod_{h \in \mathcal{H}} \phi_h)(\mathbf{c}) = \prod_{h \in \mathcal{H}} \phi_h(\mathbf{c}^{\downarrow h}),$$

where $\mathbf{c} \in v_{\mathcal{X}}$ is an arbitrary configuration and $\mathcal{X} = \cup \mathcal{H}$.

*Marginalization* :

$$\Phi_h = \begin{array}{|cccc|} \hline x_1 & \cdots & x_l & f_{\phi_h} \\ \hline c_{11} & \cdots & c_{1l} & \phi_h(c_1) \\ c_{21} & \cdots & c_{2l} & \phi_h(c_2) \\ \vdots & \vdots & \vdots & \vdots \\ c_{s1} & \cdots & c_{sl} & \phi_h(c_s) \\ \hline \end{array}$$

Figure 2: The function $\phi_h$ expressed as a relation.

Consider a function $\phi_k$ on a set $k$ of variables. If $h \subseteq k$, then $\phi_k^{\downarrow h}$ denotes the function on $h$ defined as follows:

$$\phi_k^{\downarrow h}(c_h) = \sum_{c_{k-h}} \phi_k(c_h \circ c_{k-h}),$$

where $c_h$ is a configuration of $h$, $c_{k-h}$ is a configuration of $k - h$, and $c_h \circ c_{k-h}$ is a configuration of $k$. We call $\phi_k^{\downarrow h}$ the *marginal* of $\phi_k$ on $h$.

A major task in probabilistic reasoning with belief networks is to compute marginals as new evidence becomes available.

## 3 Representation of a Factored Probability Distribution as a Generalized Acyclic Join Dependency

Let c be a configuration of $\mathcal{X} = \{x_1, x_2, ..., x_n\}$. Consider a factored probability distribution $\phi$ on $\mathcal{H}$:

$$\phi(c) = \prod_{h \in \mathcal{H}} \phi_h(c^{\downarrow h}).$$

We can conveniently express each function $\phi_h$ in the above product as a *relation* $\Phi_h$. Suppose $h = \{x_1, x_2, ..., x_l\}$. The function $\phi_h$ can be expressed as a relation on the set $\{x_1, x_2, ..., x_l, f_{\phi_h}\}$ of *attributes* as shown in Figure 2. A configuration $c_i = (c_{i1}, c_{i2}, ..., c_{il})$ in the above table denotes a row excluding the last element in the row, and $s$ is the cardinality of $v_h$.

By definition, the product $\phi_h \cdot \phi_k$ of any two function $\phi_h$ and $\phi_k$ is given by:

$$(\phi_h \cdot \phi_k)(c) = \phi_h(c^{\downarrow h}) \cdot \phi_k(c^{\downarrow k}),$$

where $c \in v_{h \cup k}$. We can therefore express the product $\phi_h \cdot \phi_k$ equivalently as a *product join* of the relations $\Phi_h$ and $\Phi_k$, written $\Phi_h \otimes \Phi_k$, which is defined as follows:

(i) Compute the *natural join*, $\Phi_h \bowtie \Phi_k$, of the two relations of $\Phi_h$ and $\Phi_k$.

(ii) Add a new column with attribute $f_{\phi_h \cdot \phi_k}$ to the relation $\Phi_h \bowtie \Phi_k$ on $h \cup k$. Each value of $f_{\phi_h \cdot \phi_k}$ is given by $\phi_h(c^{\downarrow h}) \cdot \phi_k(c^{\downarrow k})$, where $c \in v_{h \cup k}$.

(iii) Obtain the resultant relation $\Phi_h \otimes \Phi_k$ by projecting the relation obtained in Step (ii) on the set of attributes $h \cup k \cup \{f_{\phi_h \cdot \phi_k}\}$.

For example, let $h = \{x_1, x_2\}$, $k = \{x_2, x_3\}$, and $v_h = v_k = \{0, 1\}$. The product join $\Phi_h \otimes \Phi_k$ is illustrated in Figure 3.

| $x_1$ | $x_2$ | $f_{\phi_h}$ |
|---|---|---|
| 0 | 0 | $a_1$ |
| 0 | 1 | $a_2$ |
| 1 | 0 | $a_3$ |
| 1 | 1 | $a_4$ |

$\bowtie$

| $x_2$ | $x_3$ | $f_{\phi_k}$ |
|---|---|---|
| 0 | 0 | $b_1$ |
| 0 | 1 | $b_2$ |
| 1 | 0 | $b_3$ |
| 1 | 1 | $b_4$ |

(i) =

| $x_1$ | $x_2$ | $x_3$ | $f_{\phi_h}$ | $f_{\phi_k}$ |
|---|---|---|---|---|
| 0 | 0 | 0 | $a_1$ | $b_1$ |
| 0 | 0 | 1 | $a_1$ | $b_2$ |
| 0 | 1 | 0 | $a_2$ | $b_3$ |
| 0 | 1 | 1 | $a_2$ | $b_4$ |
| 1 | 0 | 0 | $a_3$ | $b_1$ |
| 1 | 0 | 1 | $a_3$ | $b_2$ |
| 1 | 1 | 0 | $a_4$ | $b_3$ |
| 1 | 1 | 1 | $a_4$ | $b_4$ |

(ii) →

| $x_1$ | $x_2$ | $x_3$ | $f_{\phi_h}$ | $f_{\phi_k}$ | $f_{\phi_h \cdot \phi_k}$ |
|---|---|---|---|---|---|
| 0 | 0 | 0 | $a_1$ | $b_1$ | $a_1 \cdot b_1$ |
| 0 | 0 | 1 | $a_1$ | $b_2$ | $a_1 \cdot b_2$ |
| 0 | 1 | 0 | $a_2$ | $b_3$ | $a_2 \cdot b_3$ |
| 0 | 1 | 1 | $a_2$ | $b_4$ | $a_2 \cdot b_4$ |
| 1 | 0 | 0 | $a_3$ | $b_1$ | $a_3 \cdot b_1$ |
| 1 | 0 | 1 | $a_3$ | $b_2$ | $a_3 \cdot b_2$ |
| 1 | 1 | 0 | $a_4$ | $b_3$ | $a_4 \cdot b_3$ |
| 1 | 1 | 1 | $a_4$ | $b_4$ | $a_4 \cdot b_4$ |

(iii) →

| $x_1$ | $x_2$ | $x_3$ | $f_{\phi_h \cdot \phi_k}$ |
|---|---|---|---|
| 0 | 0 | 0 | $a_1 \cdot b_1$ |
| 0 | 0 | 1 | $a_1 \cdot b_2$ |
| 0 | 1 | 0 | $a_2 \cdot b_3$ |
| 0 | 1 | 1 | $a_2 \cdot b_4$ |
| 1 | 0 | 0 | $a_3 \cdot b_1$ |
| 1 | 0 | 1 | $a_3 \cdot b_2$ |
| 1 | 1 | 0 | $a_4 \cdot b_3$ |
| 1 | 1 | 1 | $a_4 \cdot b_4$ |

Figure 3: The join of two relations $\Phi_h$ and $\Phi_k$.

Since the operator $\otimes$ is both commutative and associative, we can express a factored probability distribution as a join of relations:

$$\phi = \prod_{h \in \mathcal{H}} \phi_h = \bigotimes_{h \in \mathcal{H}} \Phi_h = \bigotimes \{\Phi_h | h \in \mathcal{H}\}.$$

We can also define marginalization as a relational operation. Let $\Phi_k^{\downarrow h}$ denote the relation obtained by marginalizing the function $\phi_k$ on $h \subseteq k$. We can construct the relation $\Phi_k^{\downarrow h}$ in two steps:

(a) Project the relation $\Phi_k$ on the set of attributes $h \cup \{f_{\phi_k}\}$, without eliminating identical configurations.

(b) For every configuration $c_h \in v_h$, replace the set of configurations of $h \cup \{f_{\phi_k}\}$ in the relation obtained from Step (a) by the singleton configuration $c_h \circ (\sum_{c_{k-h}} \phi_k(c_h \circ c_{k-h}))$.

$$\Phi_k = \begin{array}{|cccc|} \hline x_1 & x_2 & x_3 & f_{\phi_k} \\ \hline 0 & 0 & 0 & d_1 \\ 0 & 0 & 1 & d_2 \\ 0 & 1 & 0 & d_3 \\ 0 & 1 & 1 & d_3 \\ 1 & 0 & 0 & d_4 \\ 1 & 0 & 1 & d_4 \\ 1 & 1 & 0 & d_5 \\ 1 & 1 & 1 & d_6 \\ \hline \end{array}$$

Figure 4: A relation $\Phi_k$ with attributes $x_1, x_2, x_3, f_{\phi_k}$, and $k = \{x_1, x_2, x_3\}$.

$$\begin{array}{|ccc|} \hline x_1 & x_2 & f_{\phi_k} \\ \hline 0 & 0 & d_1 \\ 0 & 0 & d_2 \\ 0 & 1 & d_3 \\ 0 & 1 & d_3 \\ 1 & 0 & d_4 \\ 1 & 0 & d_4 \\ 1 & 1 & d_5 \\ 1 & 1 & d_6 \\ \hline \end{array}$$

Figure 5: The projection of the relation $\Phi_k$ in Figure 4 onto $\{x_1, x_2\} \cup \{f_{\phi_k}\}$.

Consider, for example, the relation $\Phi_k$ with $k = \{x_1, x_2, x_3\}$ as shown in Figure 4. Suppose we want to compute $\Phi_k^{\downarrow h}$ for $h = \{x_1, x_2\}$. From Step (a), we obtain the relation in Figure 5 by projecting $\Phi_k$ on $h \cup \{f_{\phi_k}\}$. The final result is shown in Figure 6.

Two important properties are satisfied by the operator $\downarrow$ of marginalization.

**Lemma 1**  [12, 15]

(i) If $\Phi_k$ is a relation on $k$, and $h \subseteq g \subseteq k$, then

$$(\Phi_k^{\downarrow g})^{\downarrow h} = \Phi_k^{\downarrow h}.$$

(ii) If $\Phi_h$ and $\Phi_k$ are relations on $h$ and $k$, respectively, then

$$(\Phi_h \otimes \Phi_k)^{\downarrow h} = \Phi_h \otimes (\Phi_k^{\downarrow h \cap k}).$$

Before discussing the computation of marginals of a factored distribution, let us first state the notion of computational feasibility introduced by Shafer [12]. We call a set of attributes *feasible* if it is feasible to

$$\Phi_k^{\downarrow h} = \begin{array}{|ccc|} \hline x_1 & x_2 & f_{\phi_k^{\downarrow h}} \\ \hline 0 & 0 & d_1 + d_2 \\ 0 & 1 & d_3 + d_3 \\ 1 & 0 & d_4 + d_4 \\ 1 & 1 & d_5 + d_6 \\ \hline \end{array}$$

Figure 6: The marginalization $\Phi_k^{\downarrow h}$ of the relation $\Phi_k$ in Figure 4 onto $h = \{x_1, x_2\}$.

represent relations on these attributes, join them, and marginalize on them. We assume that any subset of feasible attributes is also feasible. Furthermore, we assume that the factored distribution is represented on a hypertree and every element in $\mathcal{H}$ is feasible.

**Lemma 2**  [12, 15] Let $\Phi = \bigotimes \{\Phi_h | h \in \mathcal{H}\}$ be a factored probability distribution on a hypertree $\mathcal{H}$. Let $t$ be a twig in $\mathcal{H}$ and $b$ be a branch for $t$. Then,

(i) $(\bigotimes \{\Phi_h | h \in \mathcal{H}\})^{\downarrow \cup \mathcal{H}^{-t}}$
$= (\bigotimes \{\Phi_h | h \in \mathcal{H}^{-t}\}) \otimes \Phi_t^{\downarrow t \cap b}$.

(ii) If $k \subseteq \cup \mathcal{H}^{-t}$, then $(\bigotimes \{\Phi_h | h \in \mathcal{H}\})^{\downarrow k} = (\bigotimes \{\Phi_h^{-t} | h \in \mathcal{H}^{-t}\})^{\downarrow k}$, where $\mathcal{H}^{-t}$ denotes the set of hyperedges $\mathcal{H} - \{t\}$, $\Phi_b^{-t} = \Phi_b \otimes \Phi_t^{\downarrow t \cap b}$, and $\Phi_h^{-t} = \Phi_h$ for all other $h$ in $\mathcal{H}^{-t}$.

We now describe a procedure for computing $\Phi^{\downarrow k}$ for $k \in \mathcal{H}$, where $\Phi = \bigotimes \{\Phi_h | h \in \mathcal{H}\}$ and $\mathcal{H}$ is a hypertree. Choose a hypertree construction ordering for $\mathcal{H}$ that begins with $h_1 = k$ as the root, say $h_1, h_2, ..., h_n$, and choose a branching $b(i)$ function for this particular ordering. For $i = 1, 2, ..., n$, let

$$\mathcal{H}^i = \{h_1, h_2, ..., h_i\}.$$

This is a sequence of sub-hypertrees, each larger than the last; $\mathcal{H}^1 = \{h_1\}$ and $\mathcal{H}^n = \mathcal{H}$. The element $h_i$ is a twig in $\mathcal{H}^i$. To compute $\Phi^{\downarrow k}$, we start with $\mathcal{H}^n$ going backwards in this sequence. We use Lemma 2 each time to perform the *reduction*. At the step from $\mathcal{H}^i$ to $\mathcal{H}^{i-1}$, we go from $\Phi^{\downarrow \cup \mathcal{H}^i}$ to $\Phi^{\downarrow \cup \mathcal{H}^{i-1}}$. We omit $h_i$ in $\mathcal{H}^i$ and change the relation on $h_{b(i)}$ in $\mathcal{H}^{i-1}$ from $\Phi_{h_{b(i)}}^i$ to

$$\Phi_{h_{b(i)}}^{i-1} = \Phi_{h_{b(i)}}^i \otimes (\Phi_{h_i}^i)^{\downarrow h_i \cap h_{b(i)}},$$

and the other relations in $\mathcal{H}^{i-1}$ are not changed. The collection of relations with which we begin, $\{\Phi_h^n | h \in \mathcal{H}^n\}$, is simply $\{\Phi_h | h \in \mathcal{H}\}$, and the collection with which we end, $\{\Phi_h^1 | h \in \mathcal{H}^1\}$, consists of the single relation $\Phi_h^1 = \Phi^{\downarrow h_1}$.

Consider a factored probability distribution $\Phi = \bigotimes \{\Phi_h | h \in \mathcal{H}\}$ on a hypertree $\mathcal{H} = \{h_1, h_2, ..., h_n\}$. We say that $\Phi$ satisfies the acyclic join dependency (AJD), $*[h_1, h_2, ..., h_n]$, if $\Phi$ decomposes losslessly onto a hypertree construction ordering $h_1, h_2, ..., h_n$, i.e., $\Phi$ can be expressed as:

$$\Phi = (...((\Phi^{\downarrow h_1} \otimes' \Phi^{\downarrow h_2}) \otimes' \Phi^{\downarrow h_3})... \otimes' \Phi^{\downarrow h_n}),$$

where $\otimes'$ is a *generalized join* operator defined by:

$$\Phi^{\downarrow h_i} \otimes' \Phi^{\downarrow h_j} = \Phi^{\downarrow h_i} \otimes \Phi^{\downarrow h_j} \otimes (\Phi^{\downarrow h_i \cap h_j})^{-1}.$$

The relation $(\Phi^{\downarrow h})^{-1}$ is defined as follows. First, let us define the inverse function $(\phi^{\downarrow h})^{-1}$ of $\phi$. That is,

$$(\phi^{\downarrow h})^{-1}(\mathbf{c}) = \left(\sum_{\mathbf{c}'} \phi(\mathbf{c} \circ \mathbf{c}')\right)^{-1}, \text{ where } \sum_{\mathbf{c}'} \phi(\mathbf{c} \circ \mathbf{c}') > 0,$$

$\mathbf{c}$ is a configuration of $h \subseteq \cup \mathcal{H}$, and $\mathbf{c}'$ is a configuration of $\cup \mathcal{H} - h$. We call the function $(\phi^{\downarrow h})^{-1}$ the *inverse marginal* of $\phi$ on $h$. The *inverse relation* $(\Phi^{\downarrow h})^{-1}$ is the relation constructed from the inverse function $(\phi^{\downarrow h})^{-1}$. Obviously, the product $(\phi^{\downarrow h})^{-1} \cdot \phi^{\downarrow h}$ is a unit function on $h$, and $(\Phi^{\downarrow h})^{-1} \otimes \Phi^{\downarrow h}$ is an identity relation on $h$.

**Theorem 1** [15] *Any factored probability distribution $\Phi = \bigotimes\{\Phi_h | h \in \mathcal{H}\}$ on a hypertree, $\mathcal{H} = \{h_1, h_2, ..., h_n\}$, decomposes losslessly onto a hypertree construction ordering $h_1, h_2, ..., h_n$. That is, $\Phi$ satisfies the AJD, $*[h_1, h_2, ..., h_n]$.*

*Proof:*

Suppose $t \in \mathcal{H}$ is a twig. By Lemma 2,

$$\begin{aligned}
\Phi^{\downarrow \cup \mathcal{H}^{-t}} &= ((\bigotimes\{\Phi_h | h \in \mathcal{H}^{-t}\}) \otimes \Phi_t)^{\downarrow \cup \mathcal{H}^{-t}} \\
&= (\bigotimes\{\Phi_h | h \in \mathcal{H}^{-t}\}) \otimes \Phi_t^{\downarrow t \cap (\cup \mathcal{H}^{-t})} \\
&= (\bigotimes\{\Phi_h | h \in \mathcal{H}^{-t}\}) \otimes \Phi_t^{\downarrow t \cap b}.
\end{aligned}$$

Note that $(\Phi_t^{\downarrow t \cap b})^{-1} \otimes \Phi_t^{\downarrow t \cap b} \otimes \Phi_t = \Phi_t$, as $(\Phi_t^{\downarrow t \cap b})^{-1} \otimes \Phi_t^{\downarrow t \cap b}$ is an identity relation on $t \cap b$. Thus,

$$\begin{aligned}
&\Phi^{\downarrow \cup \mathcal{H}^{-t}} \otimes (\Phi_t^{\downarrow t \cap b})^{-1} \otimes \Phi_t \\
&= (\bigotimes\{\Phi_h | h \in \mathcal{H}^{-t}\}) \otimes \Phi_t^{\downarrow t \cap b} \otimes (\Phi_t^{\downarrow t \cap b})^{-1} \otimes \Phi_t \\
&= (\bigotimes\{\Phi_h | h \in \mathcal{H}^{-t}\}) \otimes \Phi_t \\
&= \Phi.
\end{aligned}$$

Now we want to show that:

$$(\Phi_t^{\downarrow t \cap b})^{-1} \otimes \Phi_t = (\Phi^{\downarrow t \cap b})^{-1} \otimes \Phi^{\downarrow t}.$$

Note that by property (ii) of Lemma 1, we obtain:

$$\begin{aligned}
&\Phi_t \otimes (\bigotimes\{\Phi_h | h \in \mathcal{H}^{-t}\})^{\downarrow t \cap (\cup \mathcal{H}^{-t})} \\
&= (\Phi_t \otimes (\bigotimes\{\Phi_h | h \in \mathcal{H}^{-t}\}))^{\downarrow t} \\
&= \Phi^{\downarrow t}.
\end{aligned}$$

On the other hand, we have:

$$\begin{aligned}
&(\Phi_t^{\downarrow t \cap b})^{-1} \otimes ((\bigotimes\{\Phi_h | h \in \mathcal{H}^{-t}\})^{\downarrow t \cap (\cup \mathcal{H}^{-t})})^{-1} \\
&= (\Phi_t^{\downarrow t \cap b} \otimes (\bigotimes\{\Phi_h | h \in \mathcal{H}^{-t}\})^{\downarrow t \cap (\cup \mathcal{H}^{-t})})^{-1} \\
&= ((\Phi_t \otimes (\bigotimes\{\Phi_h | h \in \mathcal{H}^{-t}\})^{\downarrow t \cap (\cup \mathcal{H}^{-t})})^{\downarrow t \cap b})^{-1} \\
&= (((\Phi_t \otimes (\bigotimes\{\Phi_h | h \in \mathcal{H}^{-t}\}))^{\downarrow t})^{\downarrow t \cap b})^{-1} \\
&= ((\Phi^{\downarrow t})^{\downarrow t \cap b})^{-1} \\
&= (\Phi^{\downarrow t \cap b})^{-1}.
\end{aligned}$$

Hence,

$$\Phi_t \otimes (\Phi_t^{\downarrow t \cap b})^{-1} = \Phi^{\downarrow t} \otimes (\Phi^{\downarrow t \cap b})^{-1}.$$

The relation $\Phi$ can therefore be expressed as:

$$\begin{aligned}
\Phi &= \Phi^{\downarrow \cup \mathcal{H}^{-t}} \otimes \Phi^{\downarrow t} \otimes (\Phi^{\downarrow (\cup \mathcal{H}^{-t}) \cap t})^{-1} \\
&= \Phi^{\downarrow \cup \mathcal{H}^{-t}} \otimes' \Phi^{\downarrow t}.
\end{aligned}$$

Moreover,

$$\begin{aligned}
\Phi^{\downarrow \cup \mathcal{H}^{-t}} &= \bigotimes\{\Phi_h | h \in \mathcal{H}^{-t}\} \otimes \Phi_t^{\downarrow t \cap b} \\
&= \bigotimes\{\Phi_h^{-t} | h \in \mathcal{H}^{-t}\}.
\end{aligned}$$

We can immediately apply the same procedure to $\Phi^{\downarrow \cup \mathcal{H}^{-t}}$ for further reduction. Thus, by applying this algorithm recursively, the desired result is obtained. □

## 4 The method for implementing a Probabilistic Inference System

In order to convert a probabilistic model into a relational model, first we need to be able to efficiently transform the input potentials into marginals. Since we assume that the hypergraph induced by the potentials is a hypertree, we can apply the propagation method [6, 12] to compute all their marginals. This process involves first moving backward along the hypertree construction ordering to find the marginal of the root, then moving forward from the root to the leaves for determining marginals of the other potentials.

The next task is to transform a probability request into a standard relational query addressed to the database which is equivalent to the original probability model. The relational query can be formulated by scanning the probability request to determine the marginals involved along the hypertree construction ordering, as well as the specific variables (attributes) within each respective marginal. Once the query is expressed in terms of the query language provided, it is then submitted to and processed by the standard database management system in the usual manner.

### 4.1 Transformation of Potentials to Marginals (Relations)

We are given as input a set of potentials $\phi_h$'s which define a factored joint probability distribution $\Phi = \bigotimes\{\Phi_h | h \in \mathcal{H}\}$, where $\mathcal{H} = \{h_1, h_2, \ldots, h_n\}$ is the corresponding hypergraph. The first step in this transformation is to check if the hypergraph $\mathcal{H}$ is a hypertree [1], but if so determine a branching function $b(i)$ for it. If we do not have a hypertree, then some potentials can be combined so that the resultant hypergraph is a hypertree [12].

In the following discussion, we henceforth assume that $\mathcal{H} = \{h_1, h_2, \ldots, h_n\}$ is a hypertree. Let the branching function $b(i), i = 2, \ldots, n$ define a hypertree construction ordering. The procedure for computing the marginal of the root $h_1$ by moving backward along the hypertree construction ordering has been described in Section 3.

Once we have determined the root marginal, $\Phi^{\downarrow h_1}$, we may move forward along the hypertree construction ordering to compute marginals of the other potentials. For this purpose, while we are moving backward we should save the relation $(\Phi_{h_i}^i)^{\downarrow h_i \cap h_{b(i)}}$ at each stage $\mathcal{H}^i$. Then we can determine the other marginals by the formula:

$$\Phi^{\downarrow h_i} = \Phi_{h_i}^i \otimes ((\Phi_{h_i}^i)^{\downarrow h_i \cap h_{b(i)}})^{-1} \otimes (\Phi^{\downarrow h_{b(i)}})^{\downarrow h_i \cap h_{b(i)}}.$$

To see this, consider the situation where we have just computed the root marginal $\Phi^{\downarrow h_1}$, namely: $\Phi^{\downarrow h_1} = \Phi_{h_1} \otimes (\Phi_{h_2}^2)^{\downarrow h_1 \cap h_2}$. Note that $\Phi_{h_2}^2 = (\Phi')^{\downarrow h_2}$, where $\Phi' = \bigotimes \{\Phi_h | h \in \mathcal{H}^{-h_1}\}$. By Lemma 1, we obtain:

$$\begin{aligned}(\Phi^{\downarrow h_1})^{\downarrow h_1 \cap h_2} &= (\Phi_{h_1} \otimes (\Phi_{h_2}^2)^{\downarrow h_1 \cap h_2})^{\downarrow h_1 \cap h_2} \\ &= \Phi_{h_1}^{\downarrow h_1 \cap h_2} \otimes (\Phi_{h_2}^2)^{\downarrow h_1 \cap h_2}.\end{aligned}$$

From Lemmas 1 and 2, it follows:

$$\begin{aligned}&\Phi_{h_2}^2 \otimes ((\Phi_{h_2}^2)^{\downarrow h_1 \cap h_2})^{-1} \otimes (\Phi^{\downarrow h_1})^{\downarrow h_1 \cap h_2} \\ &= \Phi_{h_2}^2 \otimes \Phi_{h_1}^{\downarrow h_1 \cap h_2} \\ &= (\Phi_{h_2}^2 \otimes \Phi_{h_1})^{\downarrow h_2} \\ &= \Phi^{\downarrow h_2}.\end{aligned}$$

Hence, by continuing moving forward, we will arrive at the above general formula.

Consider, for example, a factored joint probability distribution defined by six potentials [4] as shown in column 2 of Tables 1 to 6. We have modified the column names to reflect the notation used in this paper. The corresponding hypergraph, $\mathcal{H} = \{h_1 = \{x_1, x_2\}, h_2 = \{x_2, x_3, x_4, x_5\}, h_3 = \{x_2, x_4, x_5, x_6\}, h_4 = \{x_2, x_6, x_7\}, h_5 = \{x_2, x_7, x_8\}, h_6 = \{x_7, x_8, x_9\}\}$, is depicted in Figure 7. This hypergraph is in fact a hypertree. The sequence $h_1, h_2, h_3, h_4, h_5, h_6$, is a hypertree construction ordering which defines the branching function, $b(2) = 1, b(3) = 2, b(4) = 3, b(5) = 4, b(6) = 5$.

To compute the root marginal $\Phi^{\downarrow h_1}$, we may move backward from the leaf hyperedge towards the root $h_1$ along the hypertree construction ordering. Thus we first transform the hypergraph $\mathcal{H}^6(=\mathcal{H})$ to $\mathcal{H}^5$. That is, we omit $h_6$ in $\mathcal{H}^6$ and change the relation $\Phi_{h_5}$ in $\mathcal{H}^5$ to $\Phi_{h_5}^5$ defined by:

$$\begin{aligned}\Phi_{h_5}^5 &= \Phi_{h_5}^6 \otimes (\Phi_{h_6}^6)^{\downarrow h_6 \cap h_5} \\ &= \Phi_{h_5}^6 \otimes (\Phi_{h_6}^6)^{\downarrow \{x_7, x_8\}}\end{aligned}$$

and the other relations in $\mathcal{H}^5$ are not changed. Similarly we have:

$$\begin{aligned}\Phi_{h_4}^4 &= \Phi_{h_4}^5 \otimes (\Phi_{h_6}^6)^{\downarrow \{x_2, x_7\}}, \\ \Phi_{h_3}^3 &= \Phi_{h_3}^4 \otimes (\Phi_{h_6}^6)^{\downarrow \{x_2, x_6\}}, \\ \Phi_{h_2}^2 &= \Phi_{h_2}^3 \otimes (\Phi_{h_6}^6)^{\downarrow \{x_2, x_4, x_5\}}, \\ \Phi_{h_1}^1 &= \Phi_{h_1}^2 \otimes (\Phi_{h_6}^6)^{\downarrow \{x_2\}}.\end{aligned}$$

| configuration | $\phi(x_1 x_2)$ | $\Phi^{\downarrow h_1}$ | $\Phi^{\downarrow h_1}$ |
|---|---|---|---|
| $\neg x_1 \neg x_2$ | 0.502 | 0.391 | 0.391 |
| $\neg x_1\ x_2$ | 0.261 | 0.058 | 0.058 |
| $x_1 \neg x_2$ | 0.498 | 0.387 | 0.387 |
| $x_1\ x_2$ | 0.739 | 0.164 | 0.164 |

Table 1: $h_1 = \{x_1, x_2\}$

| configuration | $\phi(x_2 x_3 x_4 x_5)$ | $\Phi_{h_2}^2 \otimes ((\Phi_{h_2}^2)^{\downarrow h_2 \cap h_1})^{-1}$ | $\Phi^{\downarrow h_2}$ |
|---|---|---|---|
| $\neg x_2 \neg x_3 \neg x_4 \neg x_5$ | 0.475 | 0.569 | 0.443 |
| $\neg x_2 \neg x_3 \neg x_4\ x_5$ | 0.435 | 0.431 | 0.335 |
| $\neg x_2 \neg x_3\ x_4 \neg x_5$ | 0.000 | 0.000 | 0.000 |
| $\neg x_2 \neg x_3\ x_4\ x_5$ | 0.000 | 0.000 | 0.000 |
| $\neg x_2\ x_3 \neg x_4 \neg x_5$ | 0.000 | 0.000 | 0.000 |
| $\neg x_2\ x_3 \neg x_4\ x_5$ | 0.000 | 0.000 | 0.000 |
| $\neg x_2\ x_3\ x_4 \neg x_5$ | 0.000 | 0.000 | 0.000 |
| $\neg x_2\ x_3\ x_4\ x_5$ | 0.000 | 0.000 | 0.000 |
| $x_2 \neg x_3 \neg x_4 \neg x_5$ | 0.000 | 0.000 | 0.000 |
| $x_2 \neg x_3 \neg x_4\ x_5$ | 0.000 | 0.000 | 0.000 |
| $x_2 \neg x_3\ x_4 \neg x_5$ | 0.475 | 0.144 | 0.032 |
| $x_2 \neg x_3\ x_4\ x_5$ | 0.435 | 0.450 | 0.100 |
| $x_2\ x_3 \neg x_4 \neg x_5$ | 0.029 | 0.122 | 0.027 |
| $x_2\ x_3 \neg x_4\ x_5$ | 0.061 | 0.212 | 0.047 |
| $x_2\ x_3\ x_4 \neg x_5$ | 0.029 | 0.009 | 0.002 |
| $x_2\ x_3\ x_4\ x_5$ | 0.739 | 0.063 | 0.014 |

Table 2: $h_2 = \{x_2, x_3, x_4, x_5\}$

As $\Phi_{h_1}^1 = \Phi^{\downarrow h_1}$, we have thus determined the root marginal by moving backward. Now we start moving forward from the root. By applying the formula for computing other marginals, we immediately obtain:

$$\begin{aligned}\Phi^{\downarrow h_2} &= \Phi_{h_2}^2 \otimes ((\Phi_{h_2}^2)^{\downarrow \{x_2\}})^{-1} (\Phi^{\downarrow h_1})^{\downarrow \{x_2\}}, \\ \Phi^{\downarrow h_3} &= \Phi_{h_3}^3 \otimes ((\Phi_{h_3}^3)^{\downarrow \{x_2, x_4, x_5\}})^{-1} (\Phi^{\downarrow h_2})^{\downarrow \{x_2, x_4, x_5\}}, \\ \Phi^{\downarrow h_4} &= \Phi_{h_4}^4 \otimes ((\Phi_{h_4}^4)^{\downarrow \{x_2, x_6\}})^{-1} (\Phi^{\downarrow h_3})^{\downarrow \{x_2, x_6\}}, \\ \Phi^{\downarrow h_5} &= \Phi_{h_5}^5 \otimes ((\Phi_{h_5}^5)^{\downarrow \{x_2, x_7\}})^{-1} (\Phi^{\downarrow h_4})^{\downarrow \{x_2, x_7\}}, \\ \Phi^{\downarrow h_6} &= \Phi_{h_6}^6 \otimes ((\Phi_{h_6}^6)^{\downarrow \{x_7, x_8\}})^{-1} (\Phi^{\downarrow h_5})^{\downarrow \{x_7, x_8\}}.\end{aligned}$$

The numerical results are shown in the last column of Tables 1 to 6. These relations $\Phi^{\downarrow h_i}$ form an *acyclic join dependency* in our extended relational data model.

| configuration | $\phi(x_2 x_4 x_5 x_6)$ | $\Phi_{h_3}^3 \otimes ((\Phi_{h_3}^3)^{\downarrow h_3 \cap h_2})^{-1}$ | $\Phi^{\downarrow h_3}$ |
|---|---|---|---|
| $\neg x_2 \neg x_4 \neg x_5 \neg x_6$ | 0.561 | 0.602 | 0.267 |
| $\neg x_2 \neg x_4 \neg x_5 \;\; x_6$ | 0.371 | 0.398 | 0.176 |
| $\neg x_2 \neg x_4 \;\; x_5 \neg x_6$ | 0.519 | 0.676 | 0.226 |
| $\neg x_2 \neg x_4 \;\; x_5 \;\; x_6$ | 0.250 | 0.324 | 0.109 |
| $\neg x_2 \;\; x_4 \neg x_5 \neg x_6$ | 0.016 | 0.235 | 0.000 |
| $\neg x_2 \;\; x_4 \neg x_5 \;\; x_6$ | 0.052 | 0.765 | 0.000 |
| $\neg x_2 \;\; x_4 \;\; x_5 \neg x_6$ | 0.058 | 0.251 | 0.000 |
| $\neg x_2 \;\; x_4 \;\; x_5 \;\; x_6$ | 0.173 | 0.749 | 0.000 |
| $x_2 \neg x_4 \neg x_5 \neg x_6$ | 0.561 | 0.602 | 0.016 |
| $x_2 \neg x_4 \neg x_5 \;\; x_6$ | 0.371 | 0.398 | 0.011 |
| $x_2 \neg x_4 \;\; x_5 \neg x_6$ | 0.519 | 0.675 | 0.015 |
| $x_2 \neg x_4 \;\; x_5 \;\; x_6$ | 0.250 | 0.325 | 0.015 |
| $x_2 \;\; x_4 \neg x_5 \neg x_6$ | 0.250 | 0.235 | 0.008 |
| $x_2 \;\; x_4 \neg x_5 \;\; x_6$ | 0.052 | 0.765 | 0.026 |
| $x_2 \;\; x_4 \;\; x_5 \neg x_6$ | 0.058 | 0.251 | 0.029 |
| $x_2 \;\; x_4 \;\; x_5 \;\; x_6$ | 0.173 | 0.749 | 0.085 |

Table 3: $h_3 = \{x_2, x_4, x_5, x_6\}$

| configuration | $\phi(x_2 x_6 x_7)$ | $\Phi_{h_4}^4 \otimes ((\Phi_{h_4}^4)^{\downarrow h_4 \cap h_3})^{-1}$ | $\Phi^{\downarrow h_4}$ |
|---|---|---|---|
| $\neg x_2 \neg x_6 \neg x_7$ | 0.579 | 0.579 | 0.285 |
| $\neg x_2 \neg x_6 \;\; x_7$ | 0.421 | 0.421 | 0.208 |
| $\neg x_2 \;\; x_6 \neg x_7$ | 0.563 | 0.563 | 0.160 |
| $\neg x_2 \;\; x_6 \;\; x_7$ | 0.437 | 0.437 | 0.125 |
| $x_2 \neg x_6 \neg x_7$ | 0.579 | 0.579 | 0.049 |
| $x_2 \neg x_6 \;\; x_7$ | 0.421 | 0.421 | 0.036 |
| $x_2 \;\; x_6 \neg x_7$ | 0.563 | 0.563 | 0.077 |
| $x_2 \;\; x_6 \;\; x_7$ | 0.437 | 0.437 | 0.060 |

Table 4: $h_4 = \{x_2, x_6, x_7\}$

| configuration | $\phi(x_2 x_7 x_8)$ | $\Phi_{h_5}^5 \otimes ((\Phi_{h_5}^5)^{\downarrow h_5 \cap h_4})^{-1}$ | $\Phi^{\downarrow h_5}$ |
|---|---|---|---|
| $\neg x_2 \neg x_7 \neg x_8$ | 0.697 | 0.697 | 0.310 |
| $\neg x_2 \neg x_7 \;\; x_8$ | 0.303 | 0.303 | 0.135 |
| $\neg x_2 \;\; x_7 \neg x_8$ | 0.722 | 0.722 | 0.240 |
| $\neg x_2 \;\; x_7 \;\; x_8$ | 0.278 | 0.278 | 0.093 |
| $x_2 \neg x_7 \neg x_8$ | 0.433 | 0.433 | 0.055 |
| $x_2 \neg x_7 \;\; x_8$ | 0.567 | 0.567 | 0.071 |
| $x_2 \;\; x_7 \neg x_8$ | 0.261 | 0.261 | 0.025 |
| $x_2 \;\; x_7 \;\; x_8$ | 0.739 | 0.739 | 0.071 |

Table 5: $h_5 = \{x_2, x_7, x_8\}$

| configuration | $\phi(x_7 x_8 x_9)$ | $\Phi_{h_6}^6 \otimes ((\Phi_{h_6}^6)^{\downarrow h_6 \cap h_5})^{-1}$ | $\Phi^{\downarrow h_6}$ |
|---|---|---|---|
| $\neg x_7 \neg x_8 \neg x_9$ | 0.647 | 0.647 | 0.236 |
| $\neg x_7 \neg x_8 \;\; x_9$ | 0.353 | 0.353 | 0.129 |
| $\neg x_7 \;\; x_8 \neg x_9$ | 0.579 | 0.579 | 0.119 |
| $\neg x_7 \;\; x_8 \;\; x_9$ | 0.421 | 0.421 | 0.088 |
| $x_7 \neg x_8 \neg x_9$ | 0.594 | 0.594 | 0.157 |
| $x_7 \neg x_8 \;\; x_9$ | 0.406 | 0.406 | 0.108 |
| $x_7 \;\; x_8 \neg x_9$ | 0.438 | 0.438 | 0.072 |
| $x_7 \;\; x_8 \;\; x_9$ | 0.562 | 0.562 | 0.092 |

Table 6: $h_6 = \{x_7, x_8, x_9\}$

$$\Phi_\kappa = \Phi_{\{x_a,\ldots,x_d\}} = \begin{array}{|ccc|c|} \hline x_a & \ldots & x_d & f_{\phi_\kappa} \\ \hline c_{1a} & \ldots & c_{1d} & \phi_\kappa(\mathbf{c}_1) \\ c_{2a} & \ldots & c_{2d} & \phi_\kappa(\mathbf{c}_2) \\ \vdots & & \vdots & \vdots \\ c_{ma} & \ldots & c_{md} & \phi_\kappa(\mathbf{c}_m) \\ \hline \end{array}$$

Figure 7: The relation $\Phi_\kappa$.

## 4.2 Transformation of a Probability Request to a Query

Just as we can transform a potential $\Phi_{h_i}$ to a marginal relation $\Phi^{\downarrow h_i}$, we can transform a probability request of the form $p(x_a, \ldots, x_d | x_e = \epsilon, \ldots, x_g = \gamma)$ to a relational query. This query can then be processed by the database management system. There are, however, two ways to construct the query depending on whether the product join ($\otimes$) and generalized join ($\otimes'$) operators have been incorporated into the database management system. We will show how to transform the probability request to a relational query in either situation.

(i) In the first case we assume that the database management system has been extended to include the product join and generalized join operators. Then with respect to a particular hypertree construction ordering, we first determine the *join-path* $h_r, \ldots, h_s$ such that the union $h_r \cup \ldots \cup h_s$ of these relation schemes (hyperedges) contains all the variables in the probability request $p(x_a, \ldots, x_d | x_e = \epsilon, \ldots, x_g = \gamma)$. Then the relation $\Phi_\kappa = \Phi_{\{x_a, \ldots, x_d\}}$, depicted in Figure 7, is being constructed by the query:

    SELECT    $x_a, \ldots, x_d$
    INTO    $\Phi_\kappa$
    FROM    $\Phi^{\downarrow h_r} \otimes' \ldots \otimes' \Phi^{\downarrow h_s}$
    WHERE    $x_e = \epsilon, \ldots, x_g = \gamma$.

At this point, we have the information needed to answer the probability request all in a single relation $\Phi_\kappa$. However, to compute the required conditional probability, we need to marginalize $\Phi_\kappa$ onto $\{x_a, \ldots, x_d\}$ by the following query:

    SELECT    $x_a, \ldots, x_d, SUM(f_{\phi_\kappa})$
    INTO    $\Psi_\kappa$
    FROM    $\Phi_\kappa$
    GROUPBY    $x_a, \ldots, x_d$

Since the relation $\Psi_\kappa$ is not normalized, we have to define the normalization relation $\hat{\Psi}_\kappa$ which is a *constant* relation as shown in Figure 8, where

$$\hat{\Psi}_\kappa = \begin{array}{|cccc|} \hline x_a & \ldots & x_d & f_{\hat{\psi}_\kappa} \\ \hline c_{1a} & \ldots & c_{1d} & \hat{\psi}_\kappa(\mathbf{c}_1) = \lambda \\ c_{2a} & \ldots & c_{2d} & \hat{\psi}_\kappa(\mathbf{c}_2) = \lambda \\ \vdots & \vdots & \vdots & \vdots \\ c_{ma} & \ldots & c_{md} & \hat{\psi}_\kappa(\mathbf{c}_m) = \lambda \\ \hline \end{array}$$

Figure 8: The constant relation $\hat{\Psi}_\kappa$.

$$\lambda = \sum_{\mathbf{c}} \psi_\kappa(\mathbf{c}).$$

Finally, the answer to the probability request is given by the relation $\Psi_\kappa \otimes \hat{\Psi}_\kappa^{-1}$. This demonstrates that any probability request can be easily answered by submitting simple queries as described to the relational database management system.

The above discussion indicates that we need not implement the marginalization operator $\downarrow$, as the standard relational query languages already provide the $SUM$ and $GROUPBY$ facilities. These two functions are indeed equivalent to the marginalization operation.

(ii) In the second case we simulate the product join and generalized join operators as we are interfacing with a standard database management system. We will first discuss the simulation of the product join ($\otimes$) and generalized join ($\otimes'$) operators, before we construct the relation to answer the probability request.

Suppose we want to compute the product join of two relations $\Phi_h$ and $\Phi_k$, i.e., $\Phi_h \otimes \Phi_k$. According to the definition of $\otimes$ (see the example in Figure 3), we construct the relation $\Phi_h \bowtie \Phi_k$ by the query:

SELECT $h \cup k, f_{\phi_h}, f_{\phi_k}$
INTO $\Phi_{h \cup k}$
FROM $\Phi_h, \Phi_k$.

Next we create a new column labelled by the attribute $f_{\phi_h \cdot \phi_k}$, representing the product $\phi_h \cdot \phi_k$ by the query:

ALTER TABLE $\Phi_{h \cup k}$ ADD $f_{\phi_h \cdot \phi_k}$ FLOAT.

By definition, the entries in this column are:

$$(\phi_h \cdot \phi_k)(\mathbf{c}) = \phi_h(\mathbf{c}^{\downarrow h}) \cdot \phi_k(\mathbf{c}^{\downarrow k}),$$

where $\mathbf{c} \in v_{h \cup k}$. The following query:

UPDATE $\Phi_{h \cup k}$
SET $f_{\phi_h \cdot \phi_k} = f_{\phi_h} * f_{\phi_k}$,

accomplishes this task. The last step in simulating the product join $\otimes$ is to project $\Phi_{h \cup k}$ onto the set of attributes $h \cup k \cup \{f_{\phi_h \cdot \phi_k}\}$ using the query:

SELECT $h \cup k, f_{\phi_h \cdot \phi_k}$
INTO $\Phi_{h \otimes k}$
FROM $\Phi_{h \cup k}$.

Thus we have derived the relation $\Phi_{h \otimes k} = \Phi_h \otimes \Phi_k$.

Since $\Phi_h \otimes' \Phi_k = \Phi_h \otimes \Phi_k \otimes \Phi_{h \cap k}^{-1}$, the simulation of the generalized join $\otimes'$ is just a simple extension of the product join $\otimes$. That is we need only compute $\Phi_{h \cap k}^{-1}$, the inverse relation of $\Phi_{h \cap k}$. We construct $\Phi_{h \cap k}$ by the query:

SELECT $h \cap k, SUM(f_{\phi_h})$
INTO $\Phi_{h \cap k}$
FROM $\Phi_h$
GROUPBY $h \cap k$.

Note that we can use $SUM(f_{\phi_k})$ and $\Phi_k$ in the above query, since $\Phi_h^{\downarrow h \cap k} = \Phi_k^{\downarrow h \cap k}$. It is straightforward to construct the inverse relation $\Phi_{h \cap k}^{-1}$ from $\Phi_{h \cap k}$. Now the relation $\Phi_{h \otimes' k} = \Phi_h \otimes' \Phi_k$ is obtained by performing the product join $\Phi_{h \otimes k} \otimes \Phi_{h \cap k}^{-1}$.

Let $h_r, h_{r+1}, \ldots, h_{s-1}, h_s$ denote the join-path. We can compute the relation $\Phi_\xi = ((\ldots((\Phi^{\downarrow h_r} \otimes' \Phi^{\downarrow h_{r+1}}) \otimes' \ldots) \otimes' \Phi^{\downarrow h_{s-1}}) \otimes' \Phi^{\downarrow h_s})$ by repeatedly applying the generalized join operation. It is understood that the selection $x_e = \epsilon, \ldots, x_g = \gamma$ has been performed on each of the relations in the join-path before $\Phi_\xi$ is computed.

The relation $\Phi_\kappa$, depicted in Figure 7, is obtained by the query:

SELECT $x_a, \ldots, x_d$
INTO $\Phi_\kappa$
FROM $\Phi_\xi$.

We construct $\Psi_\kappa$ and $\hat{\Psi}_\kappa$, depicted in Figure 8, as described in (i) of this subsection. The relation $\Psi \otimes \hat{\Psi}_\kappa^{-1}$ is the answer to the given probability request.

## 5 Conclusion

Once it is acknowledged that a probabilistic model can be viewed as an extended relational data model, it immediately follows that a probabilistic model can be implemented as an everyday database application. Thus, we are spared the arduous task of designing and implementing our own probabilistic inference system and the associated costs. Even if such a system was successfully implemented, the resulting performance may not be comparable to that of existing relational databases. Our approach enables us to take advantage of the various performance enhancement techniques including query processing, query optimization, and data structure storage and manipulation, available in traditional relational database management systems. Thus the time required for belief update and answering probability requests is shortened.

The proposed relational data model also provides a unified approach to design both database and probabilistic reasoning systems.

In this paper, we have defined the product join operator $\otimes$ based on ordinary multiplication primarily because we are dealing with probabilities. By defining $\otimes$ differently (e.g. based on addition), our relational data model can be easily extended to solve a number of apparently different but closely related problems such as dynamic programming [2], solving sparse linear equations [11], and constraint propagation [3].

# References

[1] C. Beeri, R. Fagin, D. Maier and M. Yannakakis, "On the desirability of acyclic database schemes," *Journal of the Association for Computing Machinery*, vol. 30, 479–513, 1983.

[2] U. Bertelè and F. Brioschi, *Nonserial Dynamic Programming*. Academic Press, 1972.

[3] R. Dechter, A. Dechter and J. Pearl, "Optimization in constraint networks," in Influence Diagrams, Belief Nets, and Decision Analysis, edited by R.M. Oliver and J.Q. Smith, Wiley, 1990.

[4] P. Hajek, T. Havranek and R. Jirousek, *Uncertain Information Processing in Expert Systems*. CRC Press, 1992.

[5] F.V. Jensen, "Junction tree and decomposable hypergraphs," Technical report, JUDEX, Aalborg, Denmark, 1988.

[6] F.V. Jensen, S.L. Lauritzen, and K.G. Olesen, "Bayesian updating in causal probabilistic networks by local computations," *Computational Statistics Quarterly*, vol. 4, 269–282, 1990.

[7] S.L. Lauritzen and D.J. Spiegelhalterr, "Local computation with probabilities on graphical structures and their application to expert systems," *Journal of the Royal Statistical Society, Series B*, vol 50, 157–244, 1988.

[8] D. Maier, *The Theory of Relational Databases*. Computer Science Press, 1983.

[9] R.E. Neapolitan, *Probabilistic Reasoning in Expert Systems*. John Wiley and Sons, 1990.

[10] J. Pearl, *Probablistic Reasoning in Intelligent Systems: Networks of Plausible Inference*. Morgan Kaufmann, 1988.

[11] D.J. Rose, "Triangulated graphs and the elimination process," Journal of Mathematical Analysis and Its Applications, 32, 597-609, 1970.

[12] G. Shafer, "An axiomatic study of computation in hypertrees," School of Business Working Paper Series, (No. 232), University of Kansas, Lawrence, 1991.

[13] P. Shenoy, "Valuation-based systems for discrete optimization," Proc. Sixth Conference on Uncertainty in Artificial Intelligence, 334-335, 1990.

[14] S.K.M. Wong, "An extended relational data model for probablistic reasoning," submitted to publication, 1994.

[15] S.K.M. Wong, Y. Xiang and X. Nie, "Representation of bayesian networks as relational databases," Proc. Fifth International Conference Information Processing and Management of Uncertainty in Knowledge-based systems, 159-165, 1994.

[16] S.K.M. Wong, Z.W. Wang, "On axiomatization of probabilistic conditional independence," Proc. Tenth Conference on Uncertainty in Artificial Intelligence, 591-597, 1994.